\documentclass{llncs}
\usepackage{amsmath}
\usepackage{amssymb}
\usepackage{booktabs}
\usepackage{graphicx}
\usepackage{enumitem}
\usepackage{tikz}
\usetikzlibrary{positioning}

\begin{document}

\title{Formal Specification and Analysis of Autonomous Systems under Partial Compliance}

\author{Jeremy Morse\inst{1}, Dejanira Araiza-Illan\inst{1}, Jonathan Lawry\inst{2}, Arthur Richards\inst{3} and Kerstin Eder\inst{1}}

\institute{Department of Computer Science and Bristol Robotics Laboratory\\
University of Bristol, Bristol, UK\\
\email{\{jeremy.morse,dejanira.araizaillan,kerstin.eder\}@bristol.ac.uk}
\and 
Department of Engineering Mathematics\\
University of Bristol, Bristol, UK\\
\email{j.lawry@bristol.ac.uk}
\and
Department of Aerospace Engineering\\
University of Bristol, Bristol, UK\\
\email{arthur.richards@bristol.ac.uk}}

\maketitle

\begin{abstract}

The widespread adoption of autonomous systems depends on
providing guarantees of safety and functional correctness, at both
design time and runtime.
Information about the extent to which functional requirements can be met in combination with \textit{non-functional} requirements (NFRs)--i.e.\ requirements that can be partially complied with--, under dynamic and uncertain environments, provides opportunities to enhance the safety and functional correctness of systems at design time. 
We present a technique to formally define system attributes that can change or be changed to deal with dynamic and uncertain environments (denominated {\em weakened specifications}) as a partially ordered lattice, and to automatically explore the system under different specifications, using probabilistic model checking, to find the likelihood of satisfying a requirement.
The resulting probabilities form boundaries of ``optimal specifications'', analogous to Pareto frontiers in multi-objective optimization, informing the designer about the system's capabilities, such as resilience or robustness, when changing its attributes to deal with dynamic and uncertain environments.
We illustrate the proposed technique through a domestic robotic assistant example.
\end{abstract}

\section{Introduction}

Autonomous systems interact with complex uncertain and dynamic environments,
which may affect their ability to meet functional and non-functional
requirements (NFRs).
These systems should try to achieve their goals as best as
possible~\cite{Sussman2007}, regardless of the obstacles faced.
Verification and validation (V\&V) processes (e.g.\ formal analysis methods or
testing) can provide information on functional correctness and system safety, which can
be a cost-effective way to demonstrate requirements conformance at design time.

However, available system specification frameworks cannot express flexibility
in requirement compliance. It may be desirable for example, to trade a slight
``relaxation'' in an NFR or some other measure of the system (its
``attributes'') for a dramatic improvement in the
compliance of others.
Understanding the relationship between such requirements is a form of exploring
the \textit{design space} of the system, in which the designer can assess the
suitability of potential designs.
One example would be the design space for self-adaptive
systems~\cite{Brum2013}, which enumerates the set of concerns that one must
consider in an adaptive system, but for which no specification or verification
facility exists.
Formal specification languages such as temporal logics~\cite{Bellini2000},
timed input/output automata~\cite{David2010} or the Z
language~\cite{Spivey1989} require an enumeration or explicit parametrization
of all the system behaviours and their contexts (e.g.\
\cite{Gario2016,Zhao2014}), which in turn requires deep knowledge and a large
modelling effort.
More generic specification conventions such as flow charts (e.g.\ UML) or
specification documents~\cite{Hatley1988,Bowen1999} are more expressive in
terms of ``flexibility'' to requirement compliance, but they lack formal
analysis support to verify the satisfaction of requirements.

The problem of specifying NFRs in autonomous systems can be thought of as one
of feasibility: we may demand that NFRs are fully conformed with, but the
likelihood of that happening in an uncertain and dynamic environment may be
low. A desirable design is one with a high
level of NFR compliance, but that also has a high probability of achieving its objectives.
Therefore, we consider our research questions to be:
\begin{enumerate}[label={\bf RQ\arabic*.}]
\item How do we specify a system that can operate in suboptimal circumstances, allowing the system to partially (i.e.\ "as well as possible") comply with its NFRs, and how do we systematically analyse such specifications at design time? 
\item How can we assess the feasibility of a particular level of requirement
conformance, to allow comparison of designs in the design space?
\end{enumerate}

To answer the first question, we propose the use of a partially ordered lattice to define a finite set of {\em specifications} (from the system's attributes) to choose from in the design, {\em weakened} to increase the likelihood for the system to comply with functional requirements when interacting with dynamic and uncertain environments.

To answer the second question, we compute probabilities using the PRISM probabilistic model checker~\cite{Kwiatkowska2011}, according to the different specifications, based on functional requirements.  
We model the systems and their environments as parametrized Markov decision processes (MDPs), to reflect the uncertainty and dynamism of autonomous systems through non-determinism and stochasticity. 
We then use PRISM to explore the model and determine the probability, according to different specifications in the lattice, based on functional requirements. 
Traversing the lattice of specifications to find where the probabilities of requirement satisfaction increase to a desired level signifies an exploration of the possible ``relaxations'' in the system's specifications, to deal with their uncertain and dynamic environments.
Hence, our approach departs from others where design analysis imply proving that the system's assumptions are satisfied at all times, and the functional or safety and non-functional requirements are complied with always and forever.

The probabilities form a family of boundaries within the lattice elements according to transitions from $Pr<\rho$ to $Pr\geq \rho$, where $\rho$ is a threshold. 
These boundaries correspond to the ``optimal'' specifications in terms of the trade-off between high requirement compliance probability and desirable operation. 
Hence, the resulting lattice answers quantitative questions such as ``what are the specifications under which a system has to operate to successfully complete a task''.
We propose a method for the exploration of systems at design time, towards understanding the trade-off between requirement compliance and desirable operation. Our method provides means so gain confidence that autonomous systems achieve their ultimate objective when working in  dynamic, uncertain environments~\cite{Ehlers2014}.

We illustrate the use of our proposed specification technique and analysis approach through the exploration of a domestic robotic assistant, which we present alongside the theoretical aspects.
Our results show that we can analyse the behaviour of systems under dynamic, uncertain environments, gathering information about their requirement compliance without having to leave V\&V to runtime monitoring and the consequent ``failure recovery'' mechanisms.

The paper proceeds as follows. 
Section~\ref{sc:casestudy} presents a motivating case study of a domestic robotic assistant, which is used to exemplify the proposal. 
Section~\ref{sc:lattices} introduces the formalization to construct ordered system specifications.
Section~\ref{sc:modelchecking} describes the computation of probabilities according to different specifications, based on functional requirements, through probabilistic model checking. 
In Section~\ref{sc:results}, we analysed the specifications for our case study, which provided insights on ``optimal'' and preferred specifications, based on degree and likelihood of requirement compliance.
We discuss the presented approach in Section~\ref{sc:discussion}. 
Section~\ref{sc:relatedwork} presents an overview of related work on the specification and verification of systems that deal with uncertain environments at design time. 
We conclude the paper in Section~\ref{sc:conclusion} and give an outlook on future work.

\section{Motivating Case Study: A Domestic Robotic Assistant}\label{sc:casestudy}

We considered a domestic robotic assistant interacting with a person. 
This case study is representative of the adoption of autonomous systems that interact closely in the same spaces as people, and thus their behaviour needs to be verified to ensure safety and functional correctness.

\begin{figure}[t]
\centering
\includegraphics[width=0.7\columnwidth, trim=5cm 16cm 4cm 2cm, clip]{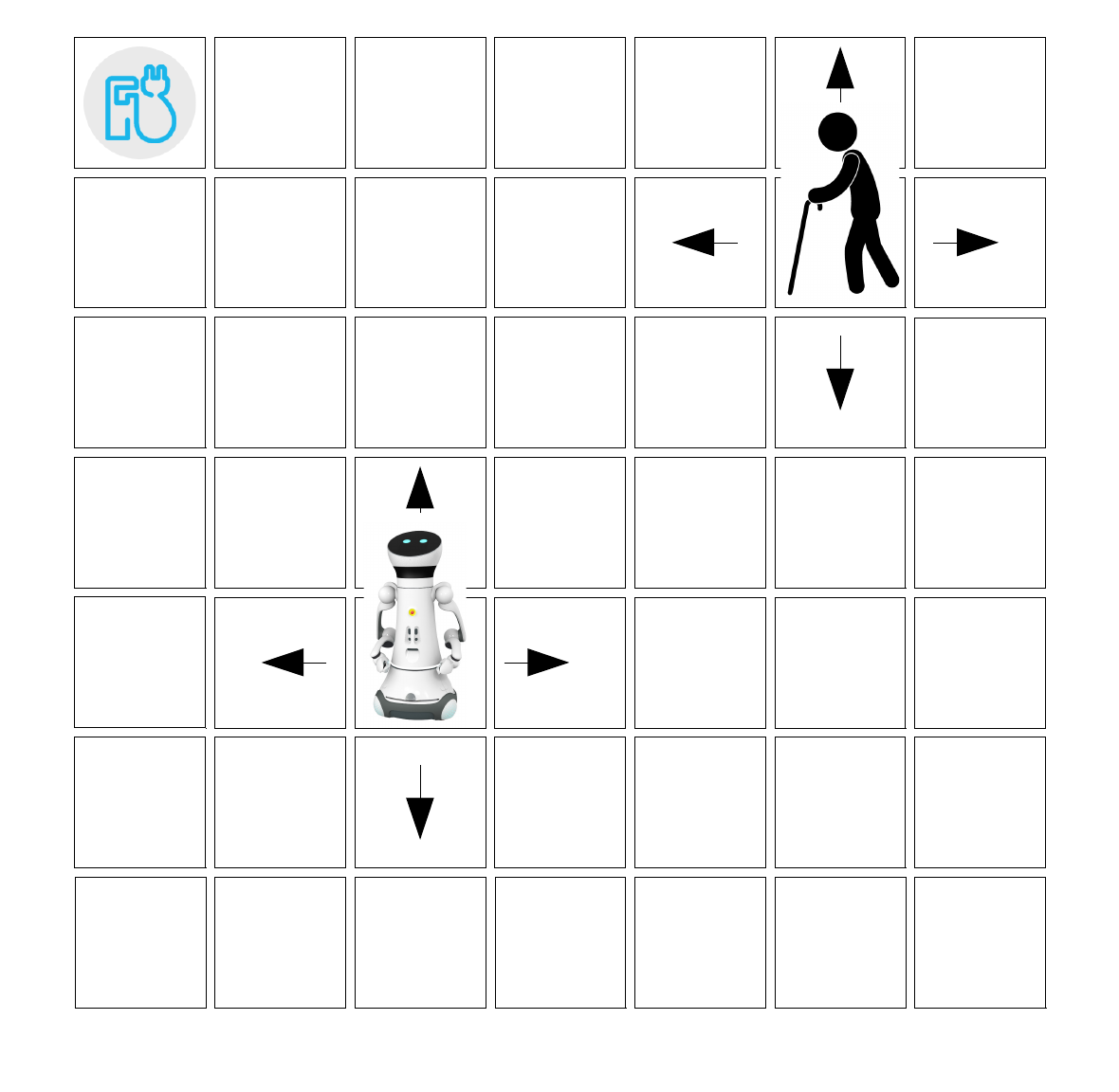} 
\caption{A robotic assistant and a person, in an open-plan, confined floor as a grid}
\label{fig:casestudy}
\end{figure}

The robot operates in an open-plan, confined floor of a house, represented by a grid. 
The robot is allowed to move in four directions, north, south, east or west in the grid, with a maximum velocity limit, or it can choose to stay in the same cell (e.g.\ to avoid collisions). 
A human that suffers of memory loss cohabits this space. This erratic behaviour is represented as stochastic motion in the grid.
A representation of this setup is shown in Fig.~\ref{fig:casestudy}.

The robot has a battery, and it will seek recharge at a station fixed in the grid, when reaching a minimal energy threshold.
The battery energy diminishes each time the robot moves (one unit per motion cycle). 
The robot needs to coordinate recharging, whilst servicing the person within a time threshold. 
Additionally, the robot avoids colliding with the person for safety. 

Due to the highly challenging human behaviour, the robot can choose to change two of its attributes: the maximum velocity limits at which it can move, and the maximum permitted servicing time. 
We would like to determine how quickly the robot can service the human with a particular maximum velocity, and what margin of energy reserves are required to complete this task.

\section{Specifying Systems that Deal with Uncertain Environments}\label{sc:lattices}

Systems that are to be verified must have a defined set of requirements that
identify the correct behaviour of the system, which a verification technique
can then prove hold over the system. 
Often, a system's requirements is augmented with a
set of NFRs, requirements that need not hold, but which are \textit{desirable} 
to hold. 
In cases where NFRs may be partially satisfied, it is desirable
they hold to the greatest extent possible. 
An example would be that a
requirement must be fulfilled in the minimum amount of time possible. 
Separately, a system may also contain configurable attributes, such as operational limits e.g.\ minimum/maximum bounds in battery charge.
We assume these attributes and the system state variables are continuous, as autonomous systems interact with the real world. 

\begin{definition}
The {\em state space} of a system is an element of $\mathbb{R}^n$.
\end{definition}
The $n$ variables of a system, e.g. time, motion, communication protocols etc,
can be related to each other. Their evolution is described by the logics and dynamics, e.g.\ an ordinary differential equation (ODE) about the system's physics or a logical description of the decision making process.
Examples of system variables in the case study are the location of the human and the robot in the grid, the battery recharge state, and the elapsed time. 

\begin{definition}
A system attribute is a function of the state variables, $f_m : \mathbb{R}^n \rightarrow Bool$, where $m$ is the attribute identifier.
\end{definition}

Each attribute constrains the state space to a subset of $\mathbb{R}^n$, $S_m \subseteq \mathbb{R}^n$, where $S_m = \{x \in \mathbb{R}^n : f_m(x) = True\}$.
Typically there is a set of attributes that, when taken together, define the states in which a system operates.
In the case study, the attributes of the robot are the maximum velocity limits, $p_i= v \leq i$ with $i=1,\ldots,6$, and the maximum allowed time cycles to service the human, $q_j = t \leq j$ with $j=1,\ldots,10$.  

\begin{definition}
Given a finite number of system attributes $f_m$, a \textit{specification} is a function $f : \mathbb{R}^n \rightarrow Bool$ such that $f = \bigwedge_m f_m$.
\end{definition}

Thus, a specification identifies the subset $S \subseteq \mathbb{R}^n$ where $S = \bigcap_m S_m$.
For the case study, this is formed by an assignment of maximum velocity limit and maximum allowed time cycles to service the human, such as $p_2= v \leq 2$ and $q_4 = t \leq 4$. 

We now consider a finite set of specifications, which can be partially ordered.
These will be used to analyse the system, to provide the system designer with information about NFR compliance and requirement satisfaction, at design time. 

Each specification for the robotic assistant has the form:
\begin{equation}\label{myspec}
	f = p_i \wedge q_j\ \mathit{where}\ i \in \{1, \ldots, 6\}\ \mathit{and}\ j \in \{1, \ldots, 10\}
\end{equation}

We also wish to be able to reason about whether a specification is ``better''
or ``worse'' than another specification, in terms of NFR satisfaction.
Therefore we define a partial order representing
the \textit{weakenings} of a specification as follows.

\begin{definition}\label{weakening}
	Given two specifications $f, f^\prime$ with associated subsets $S, S^\prime \subseteq \mathbb{R}^n$,
	we say that $f$ is a weakening of $f^\prime$, denoted $f \leq f^\prime$, iff $S \supseteq S^\prime$.
\end{definition} 

For the current example, we have that $p_i \wedge q_j \leq p_{i'} \wedge q_{j'}$ iff $i > i' \wedge j > j'$.
In other words,
considering the NFRs ``velocity must stay as small as possible'' and
``servicing time must be as fast as possible'', the \textit{strongest}
specification in the case study would be $p_1 \wedge q_1$, and the weakest
specification, i.e.\ the one with the lowest level of compliance, would be $p_6 \wedge q_{10}$.
Ideally, the robot would service the human as fast as possible (in the least possible cycles), while moving at a velocity that is not dangerous, at $p_1 \wedge q_1$. 
If this is not possible, we could settle for a ``suboptimal'' option, e.g. the specification $q_5 \wedge p_8$.
Further weakenings of the specification $p_5 \wedge q_8$ are $p_5 \wedge q_9, p_5 \wedge q_{10}, p_6 \wedge q_8, p_6 \wedge q_9$, and $p_6 \wedge q_{10}$.

\begin{figure}[t]
\centering
\begin{tikzpicture}[node distance=2cm,every node/.style={scale=0.7}]
\node(n1)                           {$p_1 \wedge q_1$};
\node(n2a)       [below =0.25cm of n1] {$p_2 \wedge q_1$};
\node(n2b)      [below right=0.25cm and 0.5cm of n1]  {$ p_1 \wedge q_2$};
\node(n3a)       [below =0.25cm of n2a] {$ p_3 \wedge  q_1$};
\node(n3b)      [below right=0.25cm and 0.5cm of n2a] {$ p_2 \wedge q_2$};
\node(n3c)      [below right=0.25cm and 2cm of n2a] {$ p_1 \wedge  q_3$};
\node(n4a)      [below =0.25cm of n3a] {$ p_4 \wedge  q_1$};
\node(n4b)      [below right=0.25cm and 0.5cm of n3a] {$ p_3 \wedge q_2$};
\node(n4c)      [below right=0.25cm and 2cm of n3a] {$ p_2 \wedge  q_3$};
\node(n4d)		[below right=0.25cm and 3.5cm of n3a] {$ p_1 \wedge  q_4$};
\node(n5a)      [below =0.25cm of n4a] {$ p_5 \wedge  q_1$};
\node(n5b)      [below right=0.25cm and 0.5cm of n4a] {$ p_4 \wedge q_2$};
\node(n5c)      [below right=0.25cm and 2cm of n4a] {$ p_3 \wedge  q_3$};
\node(n5d)		[below right=0.25cm and 3.5cm of n4a] {$ p_2 \wedge  q_4$};
\node(n5e)		[below right=0.25cm and 5cm of n4a] {$ p_1 \wedge  q_5$};
\node(n6a)      [below =0.25cm of n5a] {$ p_6 \wedge  q_1$};
\node(n6b)      [below right=0.25cm and 0.5cm of n5a] {$ p_5 \wedge q_2$};
\node(n6c)      [below right=0.25cm and 2cm of n5a] {$ p_4 \wedge  q_3$};
\node(n6d)      [below right=0.25cm and 3.5cm of n5a] {$ p_3 \wedge  q_4$};
\node(n6e)      [below right=0.25cm and 5cm of n5a] {$ p_2 \wedge  q_5$};
\node(n6f)      [below right=0.25cm and 6.5cm of n5a] {$ p_1 \wedge  q_6$};
\draw(n1)       -- (n2a);
\draw(n1) -- (n2b);
\draw(n2a) -- (n3a);
\draw(n2a) -- (n3b);
\draw(n2b) -- (n3b);
\draw(n2b) -- (n3c);
\draw(n3a) -- (n4a);
\draw(n3a) -- (n4b);
\draw(n3b) -- (n4b);
\draw(n3b) -- (n4c);
\draw(n3c) -- (n4c);
\draw(n3c) -- (n4d);
\draw(n4a) -- (n5a);
\draw(n4a) -- (n5b);
\draw(n4b) -- (n5b);
\draw(n4b) -- (n5c);
\draw(n4c) -- (n5c);
\draw(n4c) -- (n5d);
\draw(n4d) -- (n5d);
\draw(n4d) -- (n5e);
\draw(n5a) -- (n6a);
\draw(n5a) -- (n6b);
\draw(n5b) -- (n6b);
\draw(n5b) -- (n6c);
\draw(n5c) -- (n6c);
\draw(n5c) -- (n6d);
\draw(n5d) -- (n6d);
\draw(n5d) -- (n6e);
\draw(n5e) -- (n6e);
\draw(n5e) -- (n6f);
\end{tikzpicture} 
\caption{Segment of the partially ordered lattice for the domestic robotic assistant case study}
\label{fig:examplelattice}
\end{figure}
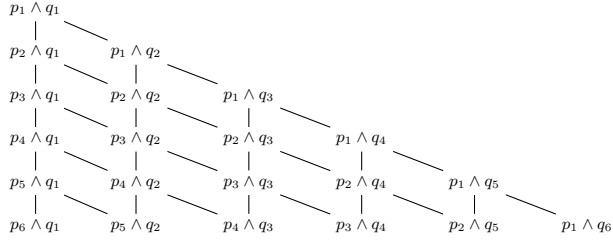

This allows the computation of a lattice using the two system attributes in the case study, and the NFRs mentioned before. 
Fig.~\ref{fig:examplelattice} shows only a segment of the resulting $6 \times 10$ lattice, built starting from the most NFR compliant specification (top).

Having formalised the system specifications to represent conformance in uncertain and dynamic environments, and ordered with respect to NFR compliance, we then proceed to explore the potential design space, to establish ``how well'' a functional requirement can be met under these specifications. 
For all specifications, given an unpredictable and dynamic environment, there are
varying probabilities of requirement satisfaction. 
For example, were one to place a low
limit on the speed of a robot, but also required it to complete tasks very
quickly, the probability of success would be low. 
This interaction between the system attributes of a specification is hard to define, and may be dependent on the
environment in which the system operates. 

Ideally, an oracle would determine the probability of the system meeting its requirements, under a given specification.
We could set a probability bound of $Pr \geq \rho$, which would separate the specifications into sets, where $Pr < \rho$ and $Pr \geq \rho$, respectively, for all possible values of $\rho$. 
We use the PRISM model checker to automatically compute these probabilities (i.e.\ as the oracle), according to different specifications, based on functional requirements.

Considering the order in the lattice, the resulting boundaries in the specifications, where the probability transitions from $Pr < \rho$ to $Pr \geq \rho$, form a surface analogous to a Pareto frontier in multi-objective optimization, representing the ``optimal'' specifications in terms of meeting the probability threshold $\rho$ at the least weakened specification with respect to NFR compliance level. In other words, the aim is to identify the maximal elements of the set of specifications with $Pr \geq \rho$.

The monotonicity of the system's NFR compliance over the state variables
translates into monotonicity of the probabilities of requirement satisfaction,
computed for the specifications in the lattice.
In the PRISM model, for any specifications $f \leq f^\prime$, by Definition~\ref{weakening}, it follows that the set of states and paths reachable for $f$ is a superset of those reachable for $f^\prime$.

The complexity to compute an entire lattice is dependent on the size of specification set.
Algorithms with time complexity of $\mathcal{O}(|X|+|\mathcal{B}|\cdot|\mathcal{B}|\cdot|\mathcal{F}|)$ have been proposed before, where $X$ and $\mathcal{B}$ are related to the size of a specification (in number of variables and samples), and $\mathcal{F}$ is the number of nodes in the lattice, formed by the conjunction or disjunction of the basis in $\mathcal{B}$ (see e.g.\ \cite{Nourine1999} for more details and algorithms). 

Although the lattice can be constructed in its entirety, a more effective exploration can be done by sampling only some of the nodes in the lattice, and applying model checking to a system under this subset of specifications to check where in the lattice the requirements are likely to be fulfilled. 
This would allow finding specific probability boundaries $Pr \leq \rho$ over the lattice in a fast and approximate manner, rather than through exploring the lattice from the top to the bottom.
We could guide the sampling and exploration of the lattice with methods used to compute approximated Pareto frontiers in multi-objective optimization, e.g.\ with Gaussian process regression~\cite{Zuluaga2012}.

\section{Systematic Analysis of System Behaviours with Probabilistic Model Checking}\label{sc:modelchecking}

In a V\&V process we determine if a system complies with its requirements. 
In classical model checking, a finite-state model of a system is explored exhaustively, to determine if a temporal logic property is satisfied or not. 
PRISM targets systems that exhibit non-deterministic behaviour, modelled as discrete-time Markov chains, continuous-time Markov chains, or Markov decision processes (MDPs), among others. 
Probabilities of property satisfaction are computed by the tool. 
The main drawback of model checking is the cost in computational time and memory, due to the state space explosion problem. 
This problem is due to the exponential growth in number of states to traverse, when adding variables~\cite{Clarke2011}. 
So far, the computational demands of model checking have made it unsuitable for runtime verification~\cite{Calinescu2010}.

We start by modelling the system and its environment as a stochastic probabilistic formal model, along with its environment.
This model is used in the PRISM probabilistic model checker, which supports verification of properties given in various temporal logics and quantifying the probability of a particular property holding.
The model is formed of three parts: the environment, the system, and a property generated from each specification based on a functional requirement.

The environment model is created as a discrete time Markov chain, where its actors are specified using probabilistic behaviour to represent uncertainty about their actions. 
All dynamic or uncertain behaviours must be encoded in the environmental model, to allow opportunity for the model checker to explore different outcomes based on those behaviours.
The system model is not encoded as a particular algorithm or controller implementation; instead all possible actions are encoded as non-deterministic choices.
This allows the model checker to pick which action the system is to take, in
any particular circumstance.
In effect, the system model represents {\em all possible} actions that
the system may choose from to respond to its environment (e.g.\ controllers), allowing the model checker to pick one that most appropriately satisfies the verification property (see below).
The combined environment and system models form a Markov decision process, or MDP.

We constructed a parametrized model of the domestic robot setting described in Section~\ref{sc:casestudy} (with a person as the stochastic environment), in the form of Markov decision processes. 
This model contains 5 \texttt{modules}, corresponding to the human and robot motion, the timing, the energy in the battery, and the servicing task. 
The parametrization prunes the state space of the model to a particular specification, which otherwise would comprise all possible specifications. 
We filter out certain unrealistic activities on the part of the robot, such as
attempting to leave the grid environment and any action that would cause an
immediate collision with the human.

The temporal logic property specification language for MDPs is probabilistic computation tree logic (PCTL)~\cite{Kwiatkowska2011}.
PRISM allows qualitative questions about the probability ($\mathcal{P}$) of properties (e.g. is is true that the probability of event $X$ is greater than $0.9$?), and also quantitative questions, i.e.\ computing actual probabilities (e.g., what is the maximum probability of event $X$, given some initial conditions?). 
The primary operator is $\phi\ U\ \psi$, ``$\phi$ holds until $\psi$ holds'',
which can be augmented with a time bound $\phi\ U_{<=t}\ \psi$ specifying a
maximum period of time $t$ in which $\psi$ must become true. Most other operators
can be derived from the until operator, which is the only operator we us in this paper.

We produce a temporal logic property from the conjunction of a
property (from functional requirements) and each specification (constraints).
Assuming that a property is already specified in PCTL, we conjoin
each state formula with the specification's constraints, producing another
PCTL formula which effectively disallows any path from entering a
state that violates the specification.
To resolve the non-determinism of the system model we replace the probability
quantification operator with the ``Pmax'' operator, specifying that
the probability should be quantified under the actions that lead to the maximum probability of
success.
The same specification in the property is also embedded in a PRISM ``filter'' statement to analyse
all starting states and select the starting state with the minimum
probability of success.
In summary, we compute the worst-case probability of success of the system
under a specification, assuming an optimal system action is
chosen from the modelled non-deterministic options.

An example of a property to verify, or a query about the model under a particular specification is: ``computing the minimum probability of completing a servicing of the human in 5 time cycles (robot motion steps), considering a specific instance of servicing time threshold bound, the maximum velocity for the robot, the minimum energy threshold, and the starting energy''.  

\footnotesize
\begin{eqnarray}\label{property}
filter(\min, 
\mathcal{P}\max=?[(serviceHuman)\wedge (velocity \leq 4) \nonumber\\ 
\mathcal{U}_{\leq 20} (\neg serviceHuman)\wedge(velocity \leq 4)], \nonumber\\
((serviceHuman \wedge serviceTimer=0) \vee \neg serviceHuman) \nonumber \\
\wedge (velocity \leq 4) \wedge (minenergy=2) \wedge (energy=25) \nonumber\\ 
\wedge (tick=0)\wedge ((robotX\neq humanX)\vee (robotY \neq humanY)))
\end{eqnarray}
\normalsize

The model checker computes a minimum probability to complete the servicing of the human, subjected to particular values of initial battery energy, maximum velocity for the robot, and maximum allowed time cycles to service the human. 
When the robot does not need to service the person, the model checker chooses a velocity and direction for its motion, from the available options, including the specified velocity upper limit. 

\section{EXPERIMENTS AND RESULTS}\label{sc:results}

The lattice was computed through a Python script. We ran the model checking experiments on a PC with Intel i5-3230M 2.60 GHz CPU,
8 GB of RAM, Ubuntu 14.04, and PRISM 4.2.beta1. 
Model checking took less than 1 minute for each experiment--i.e.\ each model checking run--, with a minimal time of 10 seconds.
All underlying data on the model and results are openly available online.\footnote{https://removed-for-review}

\begin{figure*}[t]
\centering
\includegraphics[width=\textwidth]{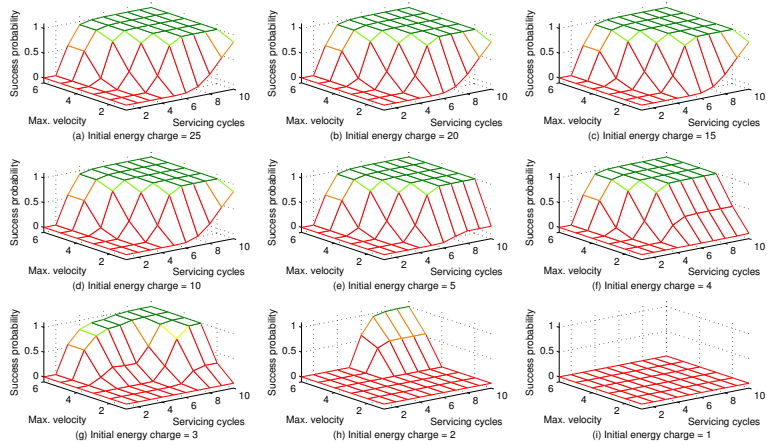} 
\caption{Probabilities for specifications in the case study, comprising the maximum velocity limit for the robot and the maximum servicing time cycles, for different initial battery energy charges}
\label{fig:graphs}
\end{figure*}

The graphs in Fig.~\ref{fig:graphs} show the probability of servicing the human, for different specifications comprising maximum velocity and maximum servicing time. 
The grid size is $7 \times 7$ cells. 
The minimum threshold for battery recharging ($min=2$ units), the initial locations of the human and the robot in the grid (at $(5,5)$ and $(4,4)$, respectively), and the location of the recharge station (at $(0,0)$ or top left corner), were left at specified fixed values. 
The initial battery energy charge was varied for each graph, from the set $\{25,20,15,10,5,4,3,2,1\}$ units, respectively.

The graphs show that the probabilities of satisfying the property~(\ref{property}) given the different specifications for the maximum velocity, maximum servicing time and available energy. The energy attributes are examined in greater detail when the initial battery charge drops below 5 units (as shown in graphs (f) to (i) in Fig.~\ref{fig:graphs}). 
We can observe that the robotic assistant will be unable to satisfy the property for the ideal condition where the velocity is minimal (i.e. $v\leq1$) and the servicing time is minimal too (i.e. $t\leq 1$), as shown in all the graphs in Fig.~\ref{fig:graphs}.
Nonetheless, we could settle for thresholds of $v\leq4$ and $t\leq 4$ (specification $p_4 \wedge q_4$), $v\leq3$ and $t\leq 5$ (specification $p_3 \wedge q_5$), or $v\leq2$ and $t\leq 6$ (specification $p_2 \wedge q_6$), if the battery is initially charged at above 5 units. 
For safety reasons, as the velocity is less, we would prefer $p_2 \wedge q_6$ over the other two options. 
Another critical threshold would be to start with a battery charged at above 4 units, since weakening the specifications to be $p_4 \wedge q_4$ or $p_3 \wedge q_5$ under this setting provides an acceptable probability of success, as shown in graph (f) in Fig.~\ref{fig:graphs}. 

\section{DISCUSSION}\label{sc:discussion}

\subsection{Answering the Research Questions}

The core contribution of our technique is providing an automated way to formalize the system specifications (through a partially ordered lattice), and to illuminate the design process towards choosing specifications (i.e.\ instances of system attributes) that satisfy their functional requirements to a high degree, whilst complying the most with respect to their NFRs. 
The obtained information about the system specifications illustrates the relations between the systems attributes. 
Fig.~\ref{fig:graphs} demonstrates these relations for our case study.

Other information that the analysis illustrated by Fig.~\ref{fig:graphs} could provide is, for example, the presence of a ``cliff edge''. 
This would show that success is almost guaranteed past a velocity limit, thus informing the designer about the little value gained by the system ever exceeding that limit. 
Likewise, if the weakening of a specification made requirement 
failure very likely (it does not ``degrade gracefully'') then a designer could
take action to avoid those specifications.

Ultimately the information imparted by our technique is limited by the
range of options available to the designer: ideally the designer would be in
a position to assign any value to any system attribute. For example, in our case
study finding that a very large amount of starting energy was required to
always complete a task might provoke the selection of a more power efficient
propulsion system. On the other hand, the system designer has increased
knowledge regarding how different levels of NFR compliance trade, and can make
appropriate decisions in the design, according
to how they prioritise different NFRs. 
Additionally, the information from the design exploration provides a set of ``optimal'' specifications, analogous to Pareto frontiers, in terms of least weakened specifications that are likely to satisfy functional requirements, whilst complying with NFRs as best as possible. 

The largely automated nature of our technique means that this information is
available to the designer without the need for her to analyse complex system
models herself. Moreover, because our probability valuations are for the
\textit{worst} starting state (given that the \textit{best} actions are chosen), the
designer is guaranteed that the reported valuation is achieved by the system
regardless of circumstance, ensuring no additional analysis is required.

Discovering and using the ``best'' system action in the model checking
process separates the concerns of which action is the best,
and whether the whole design of the system is sufficiently good. Without explicitly
considering the matter of strategy or controller synthesis, the
designer is able to discover the fundamental limitations on the performance of
their system. 

\subsection{Limitations}

While automating the exploration of a complex system to discover its
limitations is useful, it comes at the cost of having to first model the
system. Although modelling the system states might be straightforward, the
selection of appropriate probabilities with which to model the behaviour of the
environment may not be that simple, because this is likely to require either
evidence or domain knowledge. The accuracy of the resulting probability (from
the model checking process) for each specification in the lattice is
underpinned by the environmental model, and any inaccuracy there will reduce
the accuracy of the calculations.

Using a model checker for formal analysis purposes can lead to the state
explosion problem: increasing the complexity of a system model leads to an exponentially increasing number of states to be
checked. Hence, model checking does not scale gracefully. However, we see this as a
realisation of the inherent complexity in the problem we are trying to solve,
in that complex systems are difficult for humans to reason about
too. Ultimately, the matter is a cost/benefit relation: does the benefit of
modelling a particular system and the increased knowledge imparted justify the
cost of checking very large models of it?

The model checking process selects the lowest probability of success from the set of all
starting states, when calculating the probability for a particular specification.
This provides the designer with a guarantee that
the obtained probability can be met, at the cost of a result that is not
representative of all the systems' starting states. Potentially, a poor performance
for a single starting state may be acceptable to the designer, or might lead to
design changes that eliminate that starting state, but our technique does not
reveal this. One workaround for this matter would be to average the success
probability over all starting states (a feature supported by PRISM) and assess
specifications in the lattice considering both the minimum and the average probabilities.

\section{RELATED WORK}\label{sc:relatedwork}

Much of the work on the specification and verification of systems that deal with uncertain environments is based in the adaptive systems domain. The verification of adaptive systems has been performed with techniques such as formal
modelling~\cite{verifyuncertainsas,Weyns2012}, stochastic
evaluation~\cite{Rouff2012} and testing \cite{Camara2014}. These techniques
are typically applied to a particular implementation of the system's controller,
and seek to verify functional properties of the system rather than evaluate the
design space of possible controllers.  Fleurey et~al.~\cite{Fleurey2009} present a domain specific
modelling language to describe the potential adaptations of a system, and constraints to limit their context, to synthesize a requirement compliant adaptation policy. Their technique can be used to optimise policies for best
NFR compliance, but in comparison to ours, it does not fully model the environment in which the system operates. 
Their approach maximises NFR compliance for given ``tests'' (i.e.\ defined environment instances), rather than guaranteeing a level of service for different possible environments. 

Several techniques have been proposed for specifying the correct behaviour of
adaptive systems with configurable parameters, analogous to our design
space exploration but specifying the desirability of a configuration rather
than the probability of success given a system model~\cite{Cheng2009}.
Of these, RELAX~\cite{Whittle2010} provides a configuration
specification language with fuzzy logic semantics, while
FORMS~\cite{Weynsacm2012} provides a reference model for describing the correct
operation of adaptive systems.

Elsewhere, requirements engineering seeks to formulate theory and technique
behind the specification of requirements that correctly reflect a system's
behaviour, for example with adaptation \cite{requirementsprob}.
Fuzzy logic and risk analysis models can be used to formulate the system's
requirements when uncertainty is present due to
adaptation~\cite{Sawyer2010,Yang2014}. 
Qualitative requirements or soft-goals, similar to NFRs,
can also be modelled through fuzzy logic~\cite{Serrano2011}. 
In contrast,~\cite{Chopra2011} argues that the requirements should be
formulated as ``hard'' constraints (i.e. must always be fulfilled) but
contextualized (e.g., tolerances for the requirements according to the
environment), instead of fuzzy. While these techniques define a
design space, they are rarely accompanied by formal analysis and systematic design exploration methods. 

Other research on requirements engineering has focused on extracting the
requirements that are satisfied by a system.  Constraints can be removed
systematically, to provide new requirements that can be satisfied by a system,
as it is done for linear temporal logic properties in~\cite{Kim2012}. 

As well as design-time verification, adaptive systems can be subject to
runtime verification, where they are monitored to see if they satisfy
requirements or to trigger new adaptations otherwise. High level
requirements~\cite{Cheng2009}, Hidden Markov Model estimation of property
satisfaction~\cite{Bartocci2012}, and process
algebra~\cite{Abolhasanzadeh2016} have been used. Some authors propose the use
of formal methods at runtime to trigger adaptation to satisfy a requirement.  
Calinescu et al.~\cite{Calinescu2010}
use online model checking to evaluate the likelihood of
requirement satisfaction given a decision, while~\cite{onlineprism} examines
a future horizon through a model checker to determine the optimal adaptation.
While we focus on design space exploration rather than online verification,
it is conceivable that information gained through the design process can be
used to formulate better runtime strategies.

Another application of formal methods for design optimisation is synthesis of
controllers or strategies. For example, the best strategies or models of a
system that satisfy a property are computed in~\cite{Damm2011}.  Strategies
that violate a property the least are computed in~\cite{Tumova2013}, from
weighted automata and reward assignment.  These synthesis processes do not
necessarily entail adaptation (as observed in~\cite{Kressgazit2009}), as the
systems to control normally are not allowed to modify their operational spaces,
nor is the environment allowed to change dynamically, or non-deterministically. 

Ehlers et al.~\cite{Ehlers2014} use controller synthesis to search for a level
of controller resilience to ``glitches'', temporary violations of system
assumptions. Their search for an optimal system controller is similar to our
system's non-determinism, and their Pareto frontier of number-of-glitches versus
recovery-time represents our lattice to an extent. However, their exploration
is limited to only two (the aforementioned) dimensions, is not probabilistic,
and does not consider partial satisfaction of requirements (NFRs).

\section{CONCLUSIONS}\label{sc:conclusion}

We presented a technique to construct and automatically explore a specification for systems that deal with uncertain environments, for design exploration and formal analysis purposes. 
This technique constructs sections of a lattice of system specifications, partially ordered by NFR compliance. 
The lattice allows the designer to understand the trade-offs between the selection of different system attributes, in terms of increasing or decreasing functional and non-functional requirement compliance.

We illustrated the proposed technique through a domestic robotic assistant case study.
A partially ordered lattice was computed for all the attributes, as the number of specifications was small. 
Subsequently, we implemented a parametrized non-deterministic model of a robotic assistant and its home environment (including people), based on Markov decision processes (MDP) in PRISM, a probabilistic model checker.
This model encodes the world in which the robot interacts and all of its possible control policies, as well as parametrized attributes and system variables, such as minimum battery reserves and maximum speed limits.
The environment is dynamic and behaves probabilistically.
The model checker automatically computes the (minimum) probability of the system to satisfy a functional requirement (expressed as a temporal logic property) under a specification from the lattice.
Systematically exploring the lattice of specifications allowed comparing the probabilities of task success/failure, to discriminate between specifications that are suitable for the final system implementation, in a manner analogous to computing Pareto frontiers in multi-objective optimization.

Whereas for small case studies the whole lattice can be computed and explored and even visualized, partial segments of interest of a full lattice can be used instead for the analysis of real-life complex systems. 
In the near future, we will be applying this premise to a real autonomous system, for exploration and verification at design time. 
Additionally, we will research the application of our technique to runtime verification of adaptive systems.

\paragraph{\bf Acknowledgments:}

This work was supported by the EPSRC grants EP/J01205X/1, project ``RIVERAS: Robust Integrated Verification of Autonomous Systems'', and EP/K006320/1, project ``Trustworthy Robotic Assistants''.

\bibliographystyle{splncs03}

\bibliography{flexspec}

\begin{thebibliography}{10}
\providecommand{\url}[1]{\texttt{#1}}
\providecommand{\urlprefix}{URL }

\bibitem{Abolhasanzadeh2016}
Abolhasanzadeh, B., Jailili, S.: Towards modeling and runtime verification of
  self-organizing systems. Expert Systems with Applications  44,  230--244
  (2016)

\bibitem{Bartocci2012}
Bartocci, E., Grosu, R., , Karmarkar, A., Smolka, S., Stoller, S., Zadok, E.,
  Seyster, J.: Adaptive runtime verification. In: Proc. RV. pp. 168--182 (2012)

\bibitem{Bellini2000}
Bellini, P., Mattolini, R., Nesi, P.: Temporal llogic for real-time system
  specification. ACM Computing Surveys  32(1),  12--42 (2000)

\bibitem{Bowen1999}
Bowen, J., Hinchey, M.: High-integrity system specification and design.
  Springer-Verlag (1999)

\bibitem{Brum2013}
Brun, Y., Desmarais, R., Geihs, K., Litoiu, M., Lopes, A., Shaw, M., , Smit,
  M.: A design space for self-adaptive systems. In: Software Engineering for
  Self-Adaptive Systems {II} (2013)

\bibitem{Calinescu2010}
Calinescu, R., Kikuchi, S.: Formal methods at runtime. In: Proc. Monterey
  Workshops. pp. 122--135 (2010)

\bibitem{Camara2014}
C\'{a}mara, J., {de Lemos}, R., Laranjeiro, N., Ventura, R., Vieira, M.:
  Robustness evaluation of the {Rainbow} framework for self-adaptation. In:
  Proc. SAC. pp. 376--383 (2014)

\bibitem{Cheng2009}
Cheng, B., {de Lemos}, R., Giese, H., Inverardi, P., Magee, J.: Software
  engineering for self-adaptive systems: A research roadmap. Software
  Engineering for Self-Adaptive Systems pp. 1--26 (2009)

\bibitem{Chopra2011}
Chopra, A.: Requirements-driven adaptation: Compliance, context, uncertainty,
  and systems. In: Proc. REatRunTime. pp. 32--36 (2011)

\bibitem{Clarke2011}
Clarke, E., Klieber, W., Nov\'{a}\u{c}ek, M., Zuliani, P.: Model checking and
  the state explosion problem. In: Proc. LASER. pp. 1--30 (2011)

\bibitem{Damm2011}
Damm, W., Finkbeiner, B.: Does it pay to extend the perimeter of a world model?
  In: Proc. FM. pp. 12--26 (2011)

\bibitem{David2010}
David, A., Larsen, K., Legay, A., Nyman, U., Wasowski, A.: Methodologies for
  specification of real-time systems using timed {I/O} automata. In: Proc.
  FMCO. pp. 290--310 (2010)

\bibitem{Ehlers2014}
Ehlers, R., Topcu, U.: Resilience to intermittent assumption violations in
  reactive synthesis. In: Proc. HSCC (2014)

\bibitem{Fleurey2009}
Fleurey, F., Solberg, A.: A domain specific modeling language supporting
  specification, simulation and execution of dynamic adaptive systems. In:
  Proc. MODELS. pp. 606–--621 (2009)

\bibitem{Gario2016}
Gario, M., Cimatti, A., Mattarei, C., Rozier, K., Tonetta, S.: Model checking
  at scale: Automated air traffic control design space exploration. In: Proc.
  CAV (2016)

\bibitem{Hatley1988}
Hatley, D., Pirbhai, I.: Strategies For Real-Time System Specification. Dorset
  House (1988)

\bibitem{requirementsprob}
Jureta, I.J., Borgida, A., Ernst, N.A., Mylopoulos, J.: The requirements
  problem for adaptive systems. ACM Trans. Manage. Inf. Syst.  5(3),
  17:1--17:33 (Sep 2014)

\bibitem{Kim2012}
Kim, K., Fainekos, G., Sankaranarayanan, S.: On the revision problem of
  specification automata. In: Proc. ICRA. pp. 5171--5176 (2012)

\bibitem{Kressgazit2009}
{Kress-Gazit}, H., Fainekos, G., Pappas, G.: Temporal-logic-based reactive
  mission and motion planning. IEEE Transactions on Robotics  25(6),
  1370--1381 (2009)

\bibitem{Kwiatkowska2011}
Kwiatkowska, M., Norman, G., Parker, D.: {PRISM} 4.0: Verification of
  probabilistic real-time systems. In: Proc. 23rd International Conference on
  Computer Aided Verification ({CAV}'11). LNCS, vol. 6806. Springer (2011)

\bibitem{onlineprism}
Moreno, G.A., C\'{a}mara, J., Garlan, D., Schmerl, B.: Proactive
  self-adaptation under uncertainty: A probabilistic model checking approach.
  In: Proceedings of the 2015 10th Joint Meeting on Foundations of Software
  Engineering. pp. 1--12. ESEC/FSE 2015, ACM, New York, NY, USA (2015)

\bibitem{Nourine1999}
Nourine, L., Raynaud, O.: A fast algorithm for building lattices. Information
  Processing Letters  71,  199--204 (1999)

\bibitem{Rouff2012}
Rouff, C., Buskens, R., Pullum, L., Cui, X., Hinchey, M.: The {AdaptiV}
  approach to verification of adaptive systems. In: Proc. C3S2E. pp. 118--122
  (2012)

\bibitem{Sawyer2010}
Sawyer, P., Bencomo, N., Whittle, J., Leiter, E., Finkelstein, A.:
  Requirements-aware systems. In: Proc. RE. pp. 95--103 (2010)

\bibitem{Serrano2011}
Serrano, M., Serrano, M., {Sampaio do Prado Leite}, J.: Dealing with softgoals
  at runtime: A fuzzy logic approach. In: REatRunTime (2011)

\bibitem{Spivey1989}
Spivey, J.: An introduction to {Z} and formal specifications. Software
  Engineering Journal  4(1),  40--50 (1989)

\bibitem{Sussman2007}
Sussman, G.: Building robust systems an essay. In: MIT (2007)

\bibitem{Tumova2013}
Tumova, J., Castro, L., Karaman, S., Frazzoli, E., Rus, D.: Minimum-violation
  {LTL} planning with conflicting specifications. In: Proc. ACC. pp. 200--205
  (2013)

\bibitem{Weyns2012}
Weyns, D.: Towards an integrated approach for validating qualities of
  self-adaptive systems. In: Proc. WODA. pp. 24--29 (2012)

\bibitem{Weynsacm2012}
Weyns, D., Malek, S., Andersson, J.: {FORMS}: Unifying reference model for
  formal specification of distributed self-adaptive systems. ACM Transactions
  on Autonomous and Adaptive Systems  7(1) (2012)

\bibitem{Whittle2010}
Whittle, J., Sawyer, P., Bencomo, N., Cheng, B., Bruel, J.: {RELAX}: a language
  to address uncertainty in self-adaptive systems requirement. Requirements
  Engineering  15(2),  177--196 (2010)

\bibitem{verifyuncertainsas}
Yang, W., Xu, C., Liu, Y., Cao, C., Ma, X., Lu, J.: Verifying self-adaptive
  applications suffering uncertainty. In: Proceedings of the 29th ACM/IEEE
  International Conference on Automated Software Engineering. pp. 199--210. ASE
  '14, ACM, New York, NY, USA (2014),
  \url{http://doi.acm.org/10.1145/2642937.2642999}

\bibitem{Yang2014}
Yang, Z., Li, Z., Jin, Z., Chen, Y.: A systematic literature review of
  requirements modeling and analysis for self-adaptive systems. In: Proc.
  REFSQ. pp. 55--71 (2014)

\bibitem{Zhao2014}
Zhao, Y., Rozier, K.: Probabilistic model checking for comparative analysis of
  automated air traffic control systems. In: Proc. ICCAD (2014)

\bibitem{Zuluaga2012}
Zuluaga, M., Krause, A., Milder, P., P\"{u}schel, M.: ``smart'' design space
  sampling to predict {Pareto}-optimal solutions. In: Proc. LCTES. pp. 119--128
  (2012)

\end{thebibliography}

\end{document}